\documentclass[11pt]{article}

\usepackage[final]{acl}
\usepackage{xeCJK}
\usepackage{times}
\usepackage{latexsym}

\usepackage[T1]{fontenc}

\usepackage[utf8]{inputenc}

\usepackage{inconsolata}

\usepackage{graphicx}

\usepackage{url}
\usepackage{hyperref}

\usepackage[table]{xcolor}
\usepackage{amssymb}
\usepackage{amsmath}
\usepackage{tabularx}
\usepackage{array}
\usepackage{booktabs}
\usepackage{arydshln}
\usepackage{multirow}
\usepackage{multicol}
\usepackage{enumitem}
\usepackage{pifont}
\usepackage{makecell}
\usepackage[normalem]{ulem}
\usepackage{subcaption}
\usepackage[most]{tcolorbox}
\usepackage{titlesec}

\newcommand{\tabH}{\rule{0pt}{2.1ex}}
\newcommand{\bhline}{\noalign{\hrule height 1.2pt}}
\usepackage{float}
\newcommand{\mini}[1]{{\fontsize{8pt}{8pt}\selectfont #1}}
\newcommand{\after}[1]{\textcolor{black}{#1}}
\newcommand{\cameraready}[1]{\textcolor{black}{#1}}
\newcommand{\mnum}[1]{\mini{$#1\phantom{^{\text{a}}}$}}
\newcommand{\mref}[2]{\mini{$#1^{\text{#2}}$}}
\newcommand{\mna}{\mini{$-\phantom{^{\text{a}}}$}}
\newcommand{\fb}[2]{%
  \makebox[2.6em][r]{$#1$}%
  {\tiny\, $\pm$\, \makebox[2.2em][r]{$#2$}}%
}
\newcommand{\fbna}{%
  \makebox[2.6em][r]{$-$}%
  \phantom{{\tiny\, $\pm$\, \makebox[2.2em][r]{$00.0$}}}%
}
\usepackage{siunitx}
\titlespacing*{\paragraph}{0pt}{1ex}{1em}
\setlist[enumerate]{
  topsep=1pt,      
  partopsep=0pt,   
  itemsep=0.5pt,     
  parsep=0pt       
}

\title{FrameBench:\\ 
A Language Understanding Benchmark Based on Frame Semantics}

\author{%
  Chihiro Yano \hspace{2em} Ryohei Sasano\\
  Graduate School of Informatics, Nagoya University\\
  \texttt{yano.chihiro.j3@s.mail.nagoya-u.ac.jp}\quad
  \texttt{sasano@i.nagoya-u.ac.jp}
}

\begin{document}

\maketitle

\begin{abstract}
In frame semantics, sentence comprehension is assumed to proceed by relating lexical meaning to background knowledge called semantic frames, thereby enabling readers to implicitly enrich the text with unstated information. 
Recent large language models (LLMs) have achieved strong performance across a wide range of downstream tasks.
However, it remains unclear whether they can reproduce the kinds of implicit enrichment that humans naturally make during comprehension. 
To address this question, we introduce FrameBench, a benchmark grounded in frame semantics.
FrameBench consists of multiple-choice questions that test whether models distinguish the frames evoked by the same verb across contexts.
We construct the benchmark for English and Japanese using FrameNet-style resources and a generation-and-verification pipeline with native-speaker judgments.
\after{Our experiments on a diverse set of models reveal challenges for small models, while several large models}
\cameraready{surpass the human reference scores.}
We release the constructed FrameBench dataset and the code for dataset construction and evaluation at \url{https://github.com/SasanoLab/FrameBench}
\end{abstract}

\section{Introduction}
The same verb can evoke different situations depending on context. 
Consider these examples:\vspace{5pt}
\begin{enumerate}
\setlength{\itemsep}{4pt}
\item He \textit{left} the bank after talking with a friend. \label{EX::Departing}
\item He \textit{left} the bank at the age of sixty. \label{EX::Quitting}
\end{enumerate}\vspace{5pt}
\noindent Although both sentences contain the same verb, \textit{left}, they evoke different situations. 
In Sentence~\ref{EX::Departing}, \textit{left} describes a physical departure from the bank as a location. 
In Sentence~\ref{EX::Quitting}, by contrast, it evokes a quitting interpretation rather than physical departure. 
Under this interpretation, readers infer an employment relation that is not explicitly stated. 
This kind of context-dependent enrichment is central to frame semantics~\cite{Frame_Semantics}.
\begin{figure}[t]
    \centering
    \includegraphics[width=\linewidth]{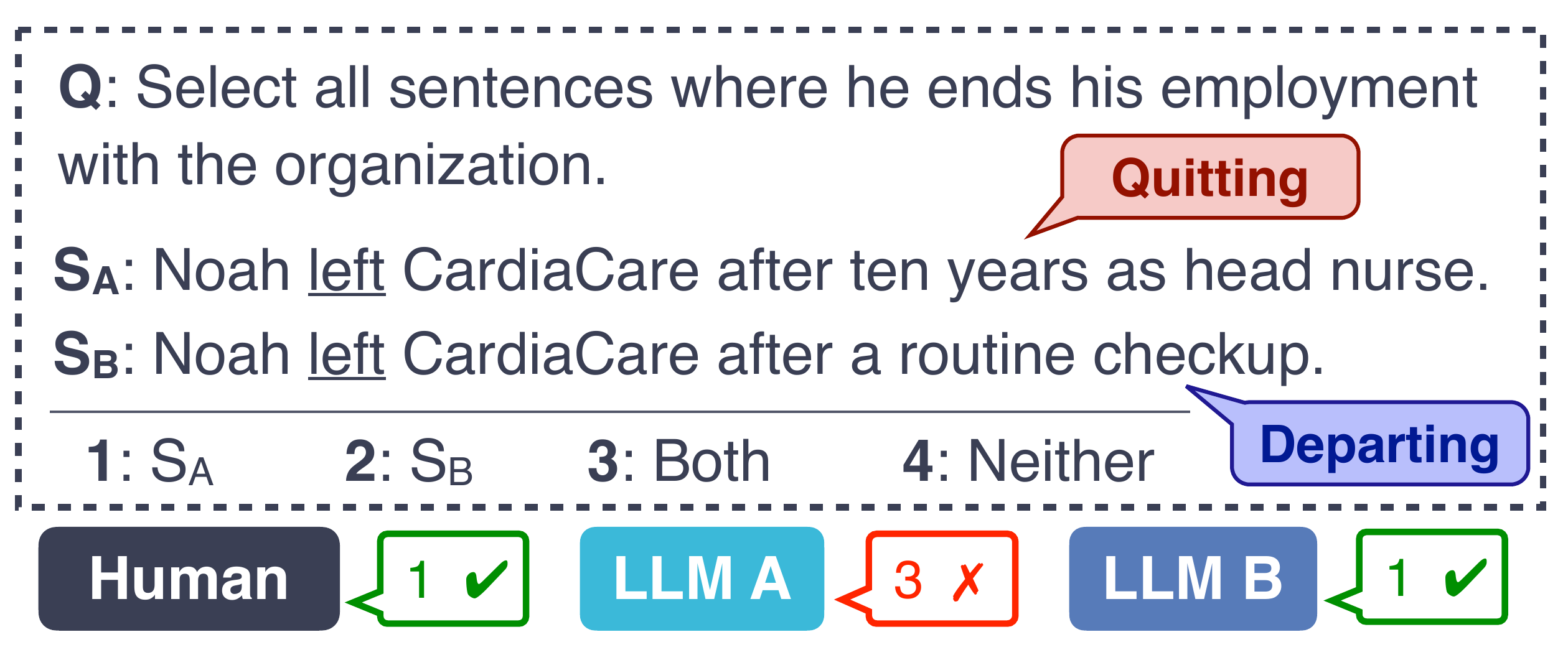}
    \caption{An example from FrameBench.}
    \label{fig:example}
\end{figure}

Despite the strong performance of recent large language models (LLMs) across a wide range of downstream tasks, it remains unclear whether they can reliably perform this kind of implicit, context-dependent enrichment.
Most existing evaluations still rely on broad benchmark suites built from diverse downstream tasks, which do not directly test this capability~\cite{glue,mmlu-pro}.
\after{To address this gap, we introduce FrameBench, a multiple-choice benchmark that evaluates whether LLMs can distinguish context-dependent frame-semantic interpretations evoked by the same verb.
Rather than requiring explicit frame-label prediction, FrameBench probes such distinctions indirectly through natural-language questions about the situations implied by each sentence.
}

In this work, we construct FrameBench for English and Japanese, two typologically distant languages, using their respective FrameNet resources \cite{FrameNet, JFN}.
\after{FrameBench is built through LLM-based generation and native-speaker validation.
Importantly, the generation process is grounded in human-authored resources rather than solely in the implicit knowledge of LLMs.}
We also evaluate a diverse set of LLMs on the resulting benchmarks to analyze whether they can distinguish the background knowledge evoked by the same verb across contexts.

Figure~\ref{fig:example} shows an example item from FrameBench.
The verb \textit{left} appears in both sentences, and the prompt asks the model to select all sentences that describe Noah quitting his job.
The correct answer is ``1: $\text{S}_\text{A}$'' because only $\text{S}_\text{A}$ expresses quitting, whereas $\text{S}_\text{B}$ describes physical departure.
The item cannot be solved by matching the predicate alone. 
It requires context-sensitive frame-semantic interpretation.

We make the following contributions:
\begin{itemize}[leftmargin=*, itemsep=2pt, parsep=0pt, topsep=0pt]

    \item \after{We introduce FrameBench, a FrameNet-grounded multiple-choice benchmark that evaluates whether LLMs can distinguish context-dependent frame-semantic interpretations evoked by the same verb.}
\item \after{
We construct English and Japanese versions of FrameBench through LLM-based generation and native-speaker validation.
We release the English and Japanese FrameBench datasets, along with the construction and evaluation code.
}
\item \after{We benchmark a wide range of LLMs, showing that FrameBench performance is strongly affected by model scale, reasoning mode, and language. Through behavioral analyses and case studies, we identify error patterns and fine-grained frame-semantic distinctions that can still challenge high-performing models.}
\end{itemize}

\section{Related Work}

\subsection{Evaluation of Language Models}
Evaluating LLMs has attracted substantial attention in recent years.
Because LLM capabilities are multifaceted, evaluations often rely on benchmarks that bundle diverse downstream tasks rather than a single task~\cite{glue, mmlu, mmlu-pro, bigbench}.

Some benchmarks more directly test whether models can distinguish meanings based on context.
For instance, Word Sense Disambiguation~(WSD) tasks and the WiC dataset~\cite{pilehvar2019wic} ask whether the same word has the same meaning in two different contexts.
Other approaches focus on sentence-level meaning representations, such as AMR~\cite{amr2017}, by evaluating how well models can parse sentences into graph-structured semantics.
However, our goal is to test whether a model can discriminate context-dependent event interpretations evoked by the same verb. 
Unlike standard WSD, which focuses on dictionary senses, our evaluation is grounded in frame semantics, capturing broader conceptual situations, which differs from the objectives of these existing resources.

\subsection{Frame-Semantic Resources and Datasets}
FrameNet is a lexical knowledge base grounded in frame semantics~\cite{Frame_Semantics}, built through manual annotation of corpus sentences with evoked frames and participant roles~\cite{FrameNet_Project}.
It has been widely used for frame-semantic parsing and related tasks such as semantic role labeling, and provides structured semantic information, including relations between frames.

Beyond English, FrameNet-style resources have been developed for many languages~\cite{JFN, korean_framenet,chinese_framenet, french_framenet, constructicography}.
More recently, datasets have been proposed that annotate multimodal data with frames and semantic roles \cite{frame2, multi40k_frame}.

\begin{figure*}
    \centering
    \includegraphics[width=1.0\linewidth]{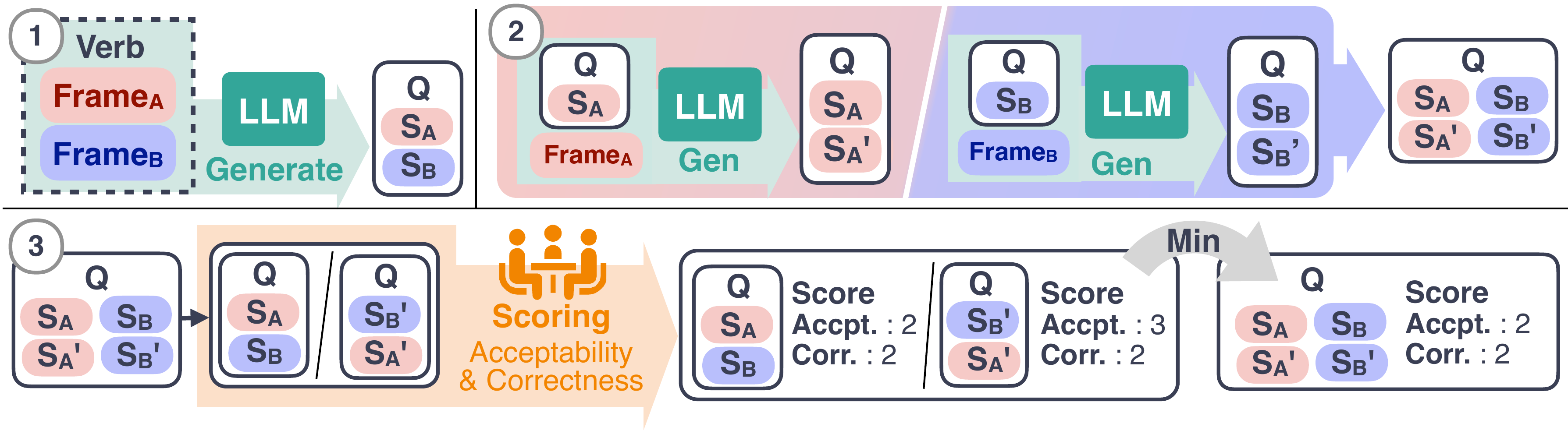}
    \caption{
    Overview of the benchmark construction. 
    The numbered steps correspond to those described in Section~\ref{sec:const}.}
    \label{fig:fig3}
\end{figure*}

\begin{figure}[t]
    \centering
    \begin{subfigure}[b]{\linewidth} 
        \includegraphics[width=0.9\linewidth]{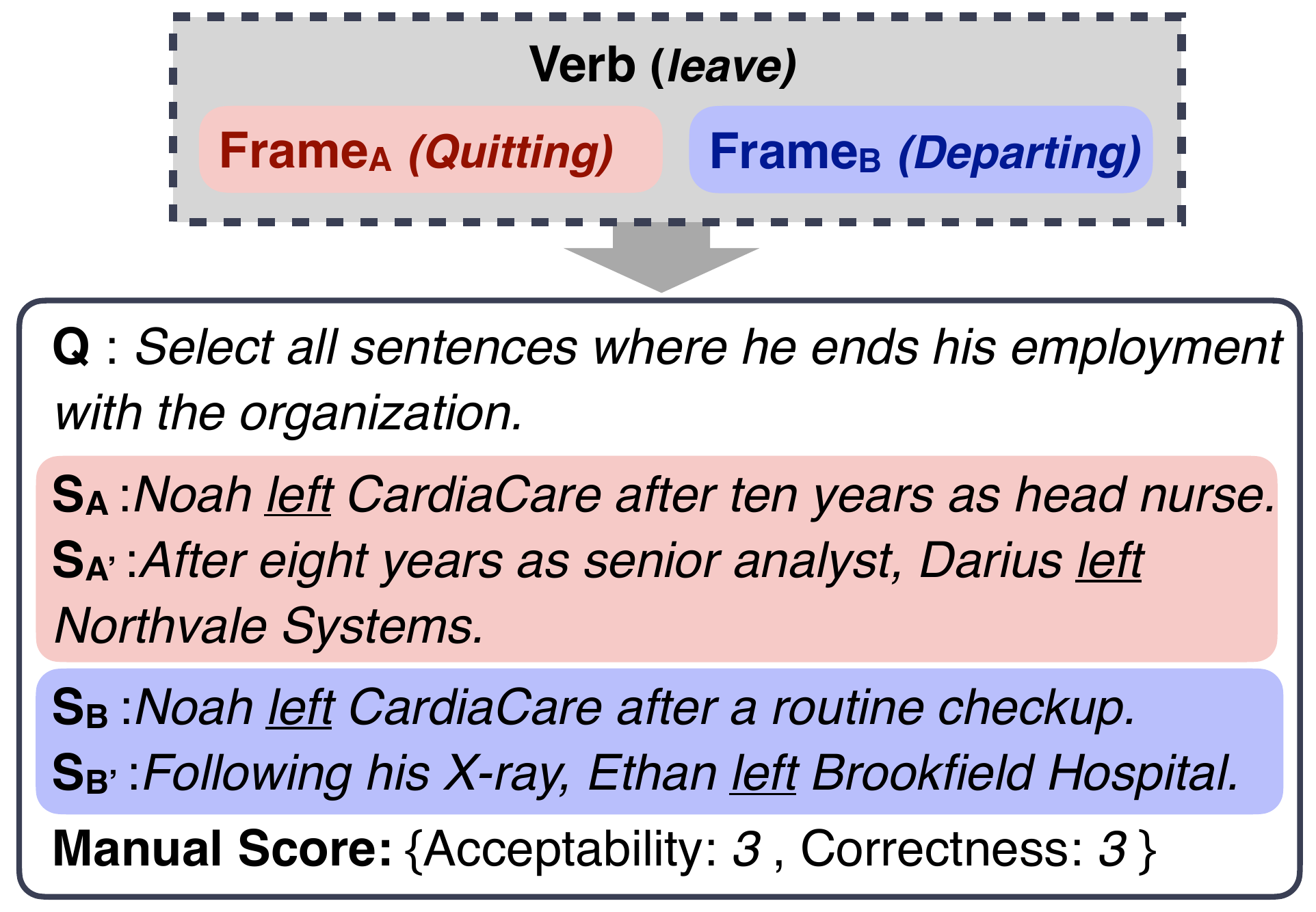}
        \caption{Example of Entry}
        \label{fig:fig2_ex_a}
    \end{subfigure}
    
    \vspace{1mm} 
    
    \begin{subfigure}[b]{\linewidth}
        \includegraphics[width=0.9\linewidth]{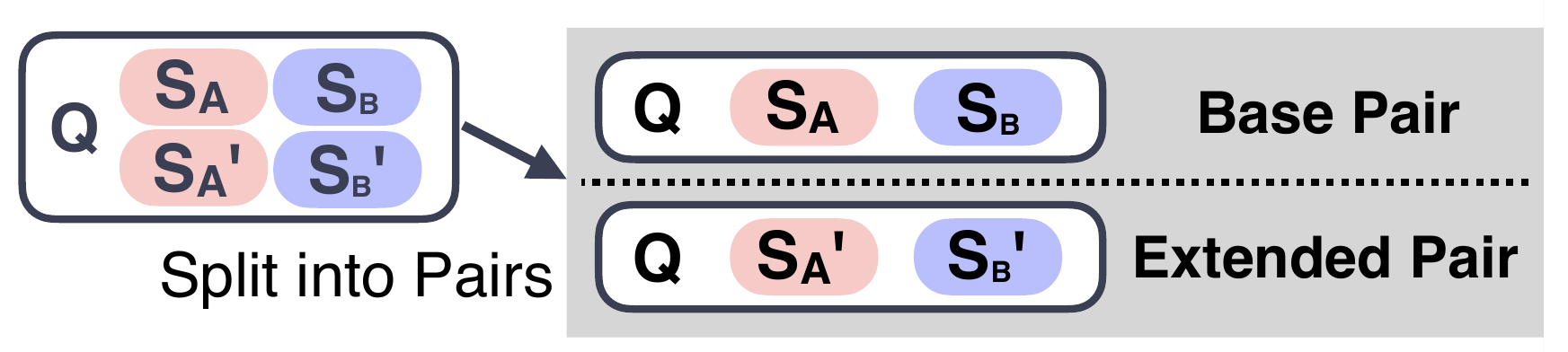}
        \caption{Evaluation Format} 
        \label{fig:fig2_ex_b}
    \end{subfigure}
    
    \caption{\after{Overview of the FrameBench task. }
    }
    \label{fig:fig2_ex}
\end{figure}
\subsection{Frame Semantics and LLMs}
A growing body of work explores the use of LLMs for frame-semantic analysis~\cite{frame_identification_llm, frame_semantic_argument_llm, icl_for_fsp,frameeol, rai-etal-2025-injecting}.
Other studies attempt to extend or assist FrameNet annotation with LLMs~\cite{llm_assited_framenet_annotation, han-etal-2024-definition}.
Overall, these results indicate that LLMs may be able to leverage frame-related information from context to some extent.

In addition, frameworks have been proposed to directly evaluate how well LLMs acquire conceptual structures grounded in frame semantics.
\citet{nutframe} propose NutFrame, which evaluates whether LLMs can induce explicit frame-semantic structures from FrameNet, and report that such induction remains challenging.
Moreover, evaluating generative or parsing tasks often suffers from formatting inconsistencies in LLM outputs.
In contrast, FrameBench evaluates the discrimination of context-dependent event interpretations evoked by the same verb as a multiple-choice task. 
By avoiding tasks that require explicit prediction of frame labels or role inventories, FrameBench more directly evaluates whether models distinguish context-dependent event interpretations evoked by the same verb.

\section{FrameBench: Task Design and Dataset Construction}
FrameBench is a four-choice benchmark for evaluating whether a model can correctly distinguish between different semantic frames evoked by the same verb.
This section first defines the FrameBench task and its evaluation protocol, then describes the construction pipeline and its application to English and Japanese.


\subsection{Task Definition}
\label{sec:task_def}
As shown in Figure~\ref{fig:fig2_ex_a}, each FrameBench entry consists of a question $Q$ and four candidate sentences ($S_{\text{A}}$, $S_{\text{B}}$, $S_{\text{A'}}$, $S_{\text{B'}}$), constructed from a polysemous verb and a pair of semantic frames ($\text{Frame}_{\text{A}}$, $\text{Frame}_{\text{B}}$).
The question $Q$ targets $\text{Frame}_{\text{A}}$:  $S_{\text{A}}$ and $S_{\text{A'}}$ are designed to evoke $\text{Frame}_{\text{A}}$, while $S_{\text{B}}$ and $S_{\text{B'}}$ are designed to evoke the contrasting $\text{Frame}_{\text{B}}$.
\after{During dataset construction, we use human-authored frame-semantic resources that specify each frame's name, definition, core frame elements, and example sentences.
Thus, the semantic distinctions in FrameBench are grounded in external frame-semantic resources rather than in an LLM's internal knowledge alone.
Since these resources are not provided at evaluation time, solving the task using only the model's internal knowledge is non-trivial.}
Each entry also includes human evaluation scores for acceptability and correctness.

\after{For evaluation, the four sentences are split into two predefined pairs, each containing exactly one sentence that evokes the target frame:
the base pair ($S_{\text{A}}$, $S_{\text{B}}$), which is constructed to be surface-similar and therefore more challenging, 
and the extended pair ($S_{\text{A'}}$, $S_{\text{B'}}$), as illustrated in Figure~\ref{fig:fig2_ex_b}.}
Given the question $Q$ and an evaluation pair, the model selects one of four labels: \textit{Sentence A}, \textit{Sentence B}, \textit{Both Sentences}, or \textit{Neither Sentence}.
\after{Although exactly one sentence is correct in each pair, we include the  two dummy options to reduce noise from forced binary choices.}

\subsection{Construction Pipeline}
\label{sec:const}
Figure~\ref{fig:fig3} provides an overview of the benchmark construction pipeline, which consists of the following three steps:

\paragraph{Step 1: Construction of Questions and \after{Base} Sentence Pairs}
\label{para:step1}
We extract polysemous verbs and their evoked frame pairs from the target language's frame-semantic resource.
For each verb $V$ and frame pair ($\text{Frame}_{\text{A}}$, $\text{Frame}_{\text{B}}$), we use an LLM to generate the base sentence pair ($S_{\text{A}}$, $S_{\text{B}}$) and a question $Q$ for which only $S_{\text{A}}$ is the correct answer.
The generation is conditioned on the frame names, definitions, core elements, and example usages of both frames.
To increase task difficulty, we instruct the LLM to make $S_{\text{A}}$ and $S_{\text{B}}$ as lexically similar as possible while preserving their distinct frame assignments.

\paragraph{Step 2: Expansion of Target Sentence Pairs}
\label{para:step2}
To diversify the evaluation set, we generate additional sentences to form the extended pair ($S_{\text{A'}}$, $S_{\text{B'}}$).
Unlike the base pair, the extended pair is not constrained to be surface-similar.
For each $\text{Frame}_\text{x}$ ($x\in\{\text{A, B}\}$), we generate an additional sentence $S_{\text{x'}}$ using the $\text{Frame}_{\text{x}}$ information, the question $Q$, and the original sentence $S_{\text{x}}$ as input.
$S_{\text{x'}}$ is constrained to evoke the same frame as $S_{\text{x}}$.
Thus, $S_{\text{x'}}$ retains the same ground-truth label as $S_{\text{x}}$ with respect to $Q$.
For example, $S_{\text{A'}}$ is a correct answer to $Q$, matching the label of $S_{\text{A}}$.


\paragraph{Step 3: Human Validation and Filtering}
\label{para:step3} 
To ensure benchmark quality, native speakers of the target language manually evaluated the constructed items. 
The evaluation consisted of two components: a correctness judgment, implemented as a four-choice task, and an acceptability judgment of the descriptions.
\after{
Model evaluation uses only single-correct pairs, whereas human evaluation additionally includes auxiliary both-correct pairs ($S_{\text{A}}, S_{\text{A'}}$) and neither-correct pairs ($S_{\text{B}}, S_{\text{B'}}$) to reduce annotator bias toward selecting a single sentence.
These auxiliary pairs are used only for validation and are excluded from entry-level scoring.
Correctness and acceptability scores are computed over the two single-correct pairs by taking the minimum across pairs for each dimension.
}

\subsection{English and Japanese Versions}
\label{sec:framebench_install}
We constructed English and Japanese versions of FrameBench.
\paragraph{Resources and Generation Setup}
As sources of frame knowledge, we used FrameNet~\cite{FrameNet} for English and Japanese FrameNet~\cite{JFN} for Japanese.
In both languages, GPT-5\footnote{Model version: 2025-08-07}~\cite{GPT5} was used for generation in Step~1 and Step~2.
Table~\ref{tab:dataset_stats} summarizes the resource and dataset statistics.
Due to differences in resource scale, we randomly sampled 800 frame pairs for English, while using all 335 eligible frame pairs for Japanese. 
For Japanese, we generated two entries per frame pair ($k{=}2$) to ensure sufficient dataset size.
During construction, we removed invalid generations, yielding 731 final items for English and 549 for Japanese.

\begin{table}[t]
\centering
\fontsize{8pt}{9pt}\selectfont
\begin{tabular}{@{\hspace{0.5em}}lrr@{\hspace{0.5em}}}
\bhline
\tabH Metric & English & Japanese \\ 
\bhline
\tabH \textit{Source FrameNet} & & \\
\# Frame Pairs ($P$)           & 800 & 335 \\
\quad \# Unique LUs          &  430   & 139 \\
\# Entries per Pair ($k$)      & 1   & 2   \\ \hline
\tabH \textit{FrameBench} & & \\
\# Candidate Items ($P \times k$)      & 800 & 670 \\
\textbf{\# Final Items}  &  \textbf{731}   & \textbf{549} \\
\quad \# Unique LUs      &  407   & 128 \\ \bhline
\end{tabular}
\caption{Statistics of the FrameBench dataset.}
\label{tab:dataset_stats}
\end{table}
\begin{table}[t]
\begin{subtable}{\linewidth}
\centering
\fontsize{8pt}{9pt}\selectfont
\vspace{-2ex}
\caption{English}
\begin{tabular}{cc|cccc|c} \bhline
& & \multicolumn{4}{c|}{\tabH \#Accpt.} & \\
& & 0 & 1 & 2 & 3 & Total\\ \hline
\multirow{4}{*}{\tabH \#Corr.} 
&\tabH 0 & 0 & 3 & 8 & 1 & 12 \\
& 1 & 2 & 4 & 15 & 10 & 31 \\
& 2 & 5 & 18 & \cellcolor{cyan!15} 48 & \cellcolor{cyan!15} 63 & 134 \\
& 3 & 11 & 55 & \cellcolor{cyan!15} 201 & \cellcolor{cyan!15} 287 & 554 \\ \hline
\multicolumn{2}{c|}{\tabH Total} & 18 & 80 & 272 & 361 & 731 \\ \bhline
\end{tabular}
\label{tab:human_eval_results_en}
\end{subtable}
\vspace{2ex}
\begin{subtable}{\linewidth}
\centering
\fontsize{8pt}{9pt}\selectfont
\caption{Japanese}
\begin{tabular}{cc|cccc|c} \bhline
& & \multicolumn{4}{c|}{\tabH \#Accpt.} & \\
& & 0 & 1 & 2 & 3 & Total\\ \hline
\multirow{4}{*}{\tabH \#Corr.} 
& \tabH 0 & 0 & 0 & 1 & 9 & 10\\
& 1 & 0 & 1 & 7 & 26 & 34 \\
& 2 & 0 & 1 & \cellcolor{cyan!15} 12 & \cellcolor{cyan!15} 117 & 130 \\
& 3 & 0 & 4 & \cellcolor{cyan!15} 34 & \cellcolor{cyan!15} 337 & 375 \\ \hline
\multicolumn{2}{c|}{\tabH Total} & 0 & 6 & 54 & 489 & 549\\ \bhline
\end{tabular}
\label{tab:human_eval_results_ja}
\end{subtable}
\vspace{-10pt}
\caption{
Distribution of entries by the numbers of annotators who judged each entry correct and acceptable.
Highlighted cells indicate the high-quality subset used for evaluation.
}
\label{tab:human_eval_results}
\end{table}

\paragraph{Manual Revision}
Compared with the English outputs, the Japanese outputs tended to contain less natural phrasing.
Therefore, one of the authors, a native Japanese speaker, manually revised the generated Japanese descriptions after Steps~1 and~2.
Low-acceptability items were further revised after the initial human validation in Step~3 and re-evaluated by annotators.

\paragraph{Dataset Statistics and Human Validation Results}
Table~\ref{tab:human_eval_results} summarizes the distribution of human validation scores for FrameBench entries.
Each entry consists of a question and two sentence pairs: the base pair and the extended pair.
Following the filtering criterion described in Step~3, we retain only entries that at least two annotators judged correct and at least two judged acceptable.
This subset defines the main evaluation set used throughout our experiments.
As shown by the blue-highlighted cells in Tables~\ref{tab:human_eval_results_en} and~\ref{tab:human_eval_results_ja}, the resulting set contains 599 English entries and 500 Japanese entries.
Detailed inter-annotator agreement statistics are provided in Appendix~\ref{app:annotator-agreement}.

\begin{table*}[ht]
\setlength{\tabcolsep}{0.75ex}
\small
\centering
\begin{tabular}{lc @{\hspace{0.5em}} c:ccc:c @{\hspace{0.5em}} c:cc:c}
\bhline
\multirow{3}{*}{\tabH Model}
& \multirow{3}{*}{$\mathcal{R}$}
& \multicolumn{5}{c}{\tabH English}
& \multicolumn{4}{c}{\tabH Japanese} \\[-0.5ex]
\cmidrule(lr){3-7}\cmidrule(lr){8-11}

& & \hspace{1.0em}\multirow{2}{*}{FrameBench} & \mini{MMLU-} & \mini{GPQA-} & \multirow{2}{*}{\mini{HLE}} & \mini{Frame} & \hspace{1.0em}\multirow{2}{*}{FrameBench}  & \mini{JamC-} & \mini{MMLU-} & \mini{Frame} \\[-0.5ex]
& &  & \mini{Pro} & \mini{Diamond} & & \mini{Ident.} &  & \mini{QA} & \mini{ProX} & \mini{Ident.} \\ \bhline

\tabH Human & - & \fb{96.3}{1.9} & \mna & - & \mna & \mna & \fb{94.9}{3.7} & \mna & \mna & \mna \\
\hline
\multicolumn{11}{l}{\rule{0pt}{2.1ex}\textbf{Closed-Source Models}} \\ \hline

\tabH GPT-5~nano & $\checkmark$ & \fb{93.5}{0.8} & \mna & \mref{67.0}{c} & \mref{7.6}{c} & \mnum{77.8} & \fb{83.9}{4.4} & \mna & \mna & \mnum{76.3} \\
GPT-5 & $\checkmark$ & \fb{99.1}{0.2} & \mref{86.5}{e} & \mref{84.2}{c} & \mref{23.5}{c} & \mnum{84.0} & \fb{96.3}{2.0} & \mref{85.8}{e} & \mref{84.9}{e} & \mnum{78.3} \\ \hline

\tabH Gemini~3.1~Flash-Lite & $\checkmark$ & \fb{97.6}{0.4} & \mref{86.2}{a} & \mref{82.2}{c} & \mref{16.2}{c} & \mnum{80.0} & \fb{97.8}{0.2} & \mna & \mna & \mnum{72.3} \\
Gemini~3.1~Pro & $\checkmark$ & \fb{99.3}{0.2} & \mref{91.2}{b} & \mref{94.1}{c} & \mref{44.7}{c} & \mnum{82.5} & \fb{97.5}{0.7} & \mna & \mna & \mnum{71.5} \\ \hline

\multicolumn{11}{l}{\rule{0pt}{2.1ex}\textbf{Open-Weight Models}} \\ \hline

\tabH Gemma~4~E2B & $\checkmark$ & \fb{86.3}{1.9} & \mref{60.0}{a} & \mref{43.3}{c} & \mref{4.8}{c} & \mnum{74.8} & \fb{82.3}{2.8} & \mref{36.1}{e} & \mref{58.0}{e} & \mnum{72.3} \\
Gemma~4~E4B & $\checkmark$ & \fb{94.5}{1.1} & \mref{69.4}{a} & \mref{57.6}{c} & \mref{3.7}{c} & \mnum{80.8} & \fb{92.2}{1.0} & \mref{42.4}{e} & \mref{67.0}{e} & \mnum{76.8} \\
Gemma~4~31B & $\checkmark$ & \fb{98.6}{0.3} & \mref{85.2}{a} & \mref{85.7}{c} & \mref{22.7}{c} & \mnum{82.0} & \fb{99.1}{0.1} & \mref{68.6}{e} & \mref{84.1}{e} & \mnum{73.8} \\ \hline

\tabH Gemma~4~E2B & - & \fb{68.6}{21.5} & \mref{57.9}{f} & \mref{40.5}{c} & \mref{4.5}{c} & \mnum{74.0} & \fb{41.5}{11.0} & \mref{33.7}{f} & \mref{50.5}{f} & \mnum{67.8} \\
Gemma~4~E4B & - & \fb{82.5}{20.1} & \mref{67.0}{f} & \mref{54.9}{c} & \mref{4.7}{c} & \mnum{71.5} & \fb{84.6}{5.1} & \mref{43.4}{f} & \mref{63.6}{f} & \mnum{68.5} \\
Gemma~4~31B & - & \fb{97.5}{0.5} & \mref{83.8}{f} & \mref{76.3}{c} & \mref{11.5}{c} & \mnum{81.0} & \fb{97.6}{0.5} & \mref{67.6}{f} & \mref{80.0}{f} & \mnum{71.5} \\ \hline

\tabH Qwen3.5-0.8B & $\checkmark$ & \fb{33.3}{7.2} & \mref{42.3}{a} & \mref{11.1}{c} & \mref{1.2}{c} & \mnum{41.0} & \fb{24.2}{1.5} & \mref{24.5}{e} & \mref{23.6}{e} & \mnum{39.3} \\
Qwen3.5-2B & $\checkmark$ & \fb{80.3}{1.1} & \mref{66.5}{a} & \mref{59.8}{c} & \mref{5.1}{c} & \mnum{68.0} & \fb{60.7}{3.3} & \mref{28.3}{e} & \mref{39.1}{e} & \mnum{54.3} \\
Qwen3.5-4B & $\checkmark$ & \fb{95.5}{0.5} & \mref{79.1}{a} & \mref{68.8}{c} & \mref{6.7}{c} & \mnum{81.0} & \fb{89.9}{2.5} & \mref{39.5}{e} & \mref{75.0}{e} & \mnum{71.0} \\
Qwen3.5-9B & $\checkmark$ & \fb{97.3}{0.2} & \mref{82.5}{a} & \mref{77.8}{c} & \mref{7.5}{c} & \mnum{81.3} & \fb{91.8}{2.4} & \mref{48.9}{e} & \mref{78.4}{e} & \mnum{72.0} \\
Qwen3.5-27B & $\checkmark$ & \fb{98.6}{0.2} & \mref{86.1}{a} & \mref{87.5}{c} & \mref{16.6}{c} & \mnum{80.8} & \fb{97.3}{0.9} & \mref{59.1}{e} & \mref{83.5}{e} & \mnum{70.8} \\ \hline

\tabH Qwen3.5-0.8B & - & \fb{47.6}{8.6} & \mref{29.7}{a} & \mref{23.6}{c} & \mref{4.9}{c} & \mnum{49.5} & \fb{25.4}{2.4} & \mref{24.0}{f} & \mref{20.9}{f} & \mnum{32.5} \\
Qwen3.5-2B & - & \fb{48.5}{11.0} & \mref{55.3}{a} & \mref{43.8}{c} & \mref{4.9}{c} & \mnum{53.3} & \fb{30.0}{3.3} & \mref{28.5}{f} & \mref{36.2}{f} & \mnum{40.0} \\
Qwen3.5-4B & - & \fb{81.1}{1.9} & \mref{69.8}{f} & \mref{71.2}{c} & \mref{7.5}{c} & \mnum{45.5} & \fb{56.7}{4.8} & \mref{37.7}{f} & \mref{60.1}{f} & \mnum{45.0} \\
Qwen3.5-9B & - & \fb{85.8}{3.2} & \mref{73.1}{f} & \mref{78.6}{c} & \mref{8.6}{c} & \mnum{64.0} & \fb{72.9}{3.9} & \mref{46.9}{f} & \mref{65.5}{f} & \mnum{47.8} \\
Qwen3.5-27B & - & \fb{96.1}{0.5} & \mref{78.3}{f} & \mref{84.2}{c} & \mref{13.2}{c} & \mnum{74.8} & \fb{83.7}{3.1} & \mref{55.9}{f} & \mref{74.5}{f} & \mnum{56.5} \\ \hline

\multicolumn{11}{l}{\rule{0pt}{2.1ex}\textbf{Japanese-Oriented Models}} \\ \hline

\tabH LLM-jp~4~8B & $\checkmark$ & \fbna & \mna & \mna & \mna & \mna & \fb{84.5}{5.8} & \mref{51.4}{e} & \mref{62.3}{e} & \mnum{75.0} \\
LLM-jp~4~8B & - & \fbna & \mna & \mna & \mna & \mna & \fb{\cameraready{69.2}}{8.8} & \mref{47.1}{f} & \mref{46.1}{f} & \mnum{33.3} \\ \hline

\tabH Qwen3~Swallow~8B & $\checkmark$ & \fbna & \mna & \mna & \mna & \mna & \fb{84.0}{4.4} & \mref{46.9}{e} & \mref{70.8}{e} & \mnum{78.8} \\
Qwen3~Swallow~32B & $\checkmark$ & \fbna & \mna & \mna & \mna & \mna & \fb{93.2}{1.1} & \mref{51.8}{e} & \mref{76.1}{e} & \mnum{76.3} \\ \hline

\multicolumn{11}{l}{\rule{0pt}{2.1ex}\textbf{Models with Shared Base Models}} \\ \hline

\tabH Qwen3-8B & $\checkmark$ & \fbna & \mna & \mna & \mna & \mna & \fb{85.0}{6.4} & \mref{40.1}{e} & \mref{71.1}{e} & \mnum{69.0} \\
Qwen3-32B & $\checkmark$ & \fbna & \mna & \mna & \mna & \mna & \fb{93.9}{0.8} & \mref{47.2}{e} & \mref{75.9}{e} & \mnum{71.3} \\ \bhline

\end{tabular}
\caption{
FrameBench results with comparison benchmark scores.
Model scores on FrameBench are reported as mean $\pm$ standard deviation over five prompt templates, while human scores are reported as mean $\pm$ standard deviation across three annotators.
$\mathcal{R}$ denotes reasoning mode, and Frame Ident. denotes the frame identification task.
Rows under ``Models with Shared Base Models'' share base models with the Qwen3~Swallow models.
For non-FrameBench scores, superscripts indicate the source of each score: 
$^a$ model-provider reports, 
$^b$~\href{https://huggingface.co/spaces/TIGER-Lab/MMLU-Pro}{MMLU-Pro leaderboard}, 
$^c$ \href{https://artificialanalysis.ai/}{Artificial Analysis}, 
$^e$ \href{https://swallow-llm.github.io/leaderboard/index-post.en.html}{Swallow LLM Leaderboard}, 
and $^f$ our own evaluations.
}
\label{tab:result_merged}
\vspace{-0.5ex}
\end{table*}

\section{Evaluation on FrameBench}
\label{sec:experiment}
We evaluate LLMs on the English and Japanese versions of FrameBench and compare their performance with existing benchmarks.

\subsection{Experimental Setup}
\after{
We evaluate models on the English and Japanese main evaluation sets described in Section~\ref{sec:framebench_install}.
For each entry, we evaluate both the base pair and the extended pair.
To mitigate position bias, each pair is tested in both original and swapped orders.
We report accuracy as the mean and standard deviation across five prompt templates per language.
Answer options are randomly ordered for each question.
}

\after{We test closed-source, open-weight, and Japanese-oriented models, using constrained or structured decoding for reliable answer extraction. 
As the human score, we report accuracy computed from the human validation results described in Section~\ref{sec:framebench_install}. 
Since the evaluation subset is filtered to entries answered correctly by at least two annotators, the human score may be upwardly biased.
Note that an independently measured human score may be lower if the same task were administered in a separate evaluation round.
}

\after{
For comparison, we report results on widely used LLM evaluation benchmarks: MMLU-Pro~\cite{mmlu-pro}, GPQA-Diamond~\cite{gpqa}, Humanity's Last Exam~(HLE)~\cite{hle}, JamC-QA~\cite{jamcqa}, and MMLU-ProX~\cite{mmluprox}.
We additionally evaluate a frame identification task as a complementary measure of models' ability to identify frames from sentences.
Additional details of the experimental setup are provided in Appendix~\ref{app:experimental_details}.
}

\subsection{Results}
\label{subsec:exp_en}
\after{
Table~\ref{tab:result_merged} shows the results of the English and Japanese evaluations.
Across both languages, FrameBench performance increases with model scale and reasoning.
Several high-performing models surpass the human reference scores, despite the potential upward bias in the human scores.
The scale and reasoning trends are broadly consistent with the results on the comparison benchmarks.
However, reasoning gains tend to be larger on FrameBench than on several of these benchmarks.
}

\after{
We first focus on the English results.
Among closed-source models, Gemini 3.1 Pro achieves the highest score, reaching 99.3, followed by GPT-5 at 99.1.
Among open-weight models, Qwen3.5-27B and Gemma~4~31B perform best, both reaching 98.6 with reasoning.
All of these scores exceed the human reference score of 96.3.
Across the results, larger models consistently obtain higher scores, and reasoning improves performance for all matched open-weight models except Qwen3.5-0.8B.
The gains are particularly large for smaller and mid-sized models.
For example, enabling reasoning improves Qwen3.5-2B from 48.5 to 80.3 and Qwen3.5-4B from 81.1 to 95.5.}

\after{
Compared with existing LLM benchmarks, FrameBench shows broadly consistent model rankings.
Frame identification scores also show a broadly similar trend to FrameBench, but with several reversals relative to the comparison benchmarks.
Together, these patterns suggest that FrameBench reflects general language-understanding ability while remaining grounded in frame-semantic interpretation.
This makes FrameBench a more suitable evaluation setting than direct frame identification for assessing frame-semantic language understanding in LLMs.
At the same time, FrameBench shows larger performance gains than the comparison benchmarks as model capability improves through scaling or reasoning.
For example, Qwen3.5 with reasoning improves from 33.3 at 0.8B to 80.3 at 2B in English, a much larger jump than the corresponding increase on MMLU-Pro.
This pattern suggests that FrameBench is particularly sensitive to the ability to discriminate context-dependent frame-semantic interpretations.
}


\cameraready{
The LLM used in dataset construction could affect the resulting FrameBench entries and the evaluation results.
To examine this possibility, we constructed an additional 100 English FrameBench entries using Gemini 3.1 Pro in place of GPT-5 and evaluated them.
The overall performance pattern was similar to that observed on the original FrameBench.
Detailed results are provided in Appendix~\ref{app:generation_model}.}

\after{
Turning to the Japanese results, we observe similar trends, with performance generally improving with model scale and reasoning.
This is in line with the results on MMLU-ProX.
Gemma~4~31B achieves the highest score of 99.1, followed by Gemini~3.1~Flash-Lite at 97.8, Gemini~3.1~Pro at 97.5, Qwen3.5-27B at 97.3, and GPT-5 at 96.3.
All of these scores exceed the human reference score of 94.9.
Reasoning also yields sharp gains, improving Qwen3.5-2B from 30.0 to 60.7, Qwen3.5-4B from 56.7 to 89.9, Gemma~4~E2B from 41.5 to 82.3, and Gemma~4~E4B from 84.6 to 92.2.
}

\after{For Japanese-oriented models, we compare Qwen3~Swallow models with the corresponding Qwen3 models, since they are based on the same Qwen3-Base models.
Although the Swallow models outperform their Qwen3 counterparts on JamC-QA, they perform comparably to or slightly below the corresponding Qwen3 models on FrameBench.
Specifically, Qwen3~Swallow~8B scores 84.0 compared with 85.0 for Qwen3-8B, and Qwen3~Swallow~32B scores 93.2 compared with 93.9 for Qwen3-32B.
This contrast suggests that target-language training does not necessarily improve frame-semantic interpretation.
}

\after{Comparing the English and Japanese results, the human reference score decreases only slightly, from 96.3 to 94.9, whereas many LLMs show larger drops.
Moreover, the Japanese results separate models more clearly than the English results.
For example, strong reasoning models show small cross-lingual gaps: Gemma~4~31B changes only from 98.6 to 99.1, and Qwen3.5-27B from 98.6 to 97.3.
In contrast, weaker models show larger drops, with GPT-5~nano falling from 93.5 to 83.9 and Qwen3.5-9B without reasoning from 85.8 to 72.9.
These results suggest that Japanese FrameBench cannot be solved by the cross-lingual transfer ability of lower-capability models alone, and instead requires robust context-dependent semantic interpretation in Japanese.}

\begin{table*}[t!]
\centering
\fontsize{8pt}{9pt}\selectfont
\begin{tabular}{
  l c
  S[table-format=2.1]
  S[table-format=2.1] S[table-format=2.1] S[table-format=-1.1]
  p{0.05cm}
  S[table-format=2.1] S[table-format=2.1] S[table-format=+2.1]
  p{0.05cm}
  S[table-format=2.1] S[table-format=2.1] S[table-format=2.1]
}
\bhline
&&& \multicolumn{3}{c}{\tabH (i) Sent. Pair Type} & & \multicolumn{3}{c}{\tabH (ii) Correct Sent. Position} & & \multicolumn{3}{c}{\tabH (iii) Error Breakdown} \\ 
\cline{4-6} \cline{8-10} \cline{12-14}
\tabH Model & $\mathcal{R}$ & \multicolumn{1}{c}{\tabH Acc.} & \multicolumn{1}{c}{\tabH Base} & \multicolumn{1}{c}{\tabH Ext.} &  \multicolumn{1}{c}{\tabH $\Delta_{B-E}$} & & \multicolumn{1}{c}{\tabH 1st} & \multicolumn{1}{c}{\tabH 2nd} &  \multicolumn{1}{c}{\tabH $\Delta_{1-2}$} & & \multicolumn{1}{c}{\tabH Both} & \multicolumn{1}{c}{\tabH Neither} & \multicolumn{1}{c}{\tabH Opposite} \\ \bhline
\tabH Gemma~4~E2B & - & 68.6 & 66.9 & {\bfseries 70.3} & -3.3 & & {\bfseries 77.1} & 60.1 & +17.0 & & 11.7 & 7.7 & {\bfseries 12.0} \\
Gemma~4~E4B & - & 82.5 & 80.8 & {\bfseries 84.1} & -3.3 & & {\bfseries 84.2} & 80.8 & +3.4 & & 6.3 & 4.6 & {\bfseries 6.6} \\
Gemma~4~31B & - & 97.5 & 96.8 & {\bfseries 98.2} & -1.4 & & {\bfseries 97.5} & {\bfseries 97.5} & 0.0 & & {\bfseries 2.2} & 0.2 & 0.1 \\
\hline
\tabH Gemma~4~E2B & \checkmark & 86.3 & 85.4 & {\bfseries 87.1} & -1.7 & & {\bfseries 89.2} & 83.4 & +5.7 & & {\bfseries 10.8} & 2.1 & 0.8 \\
Gemma~4~E4B & \checkmark & 94.5 & 93.8 & {\bfseries 95.1} & -1.4 & & {\bfseries 95.6} & 93.4 & +2.2 & & {\bfseries 4.6} & 0.6 & 0.3 \\
Gemma~4~31B & \checkmark & 98.6 & 98.2 & {\bfseries 99.1} & -0.9 & & {\bfseries 98.9} & 98.3 & +0.6 & & {\bfseries 1.1} & 0.1 & 0.2 \\
\hline
\tabH Qwen3.5-0.8B & - & 47.6 & 46.2 & {\bfseries 49.0} & -2.7 & & {\bfseries 52.9} & 42.4 & +10.5 & & 24.1 & 2.9 & {\bfseries 25.4} \\
Qwen3.5-2B & - & 48.5 & 46.4 & {\bfseries 50.5} & -4.2 & & {\bfseries 54.4} & 42.6 & +11.9 & & 15.3 & {\bfseries 22.5} & 13.7 \\
Qwen3.5-4B & - & 81.1 & 80.5 & {\bfseries 81.6} & -1.1 & & {\bfseries 90.8} & 71.4 & +19.4 & & {\bfseries 7.9} & 4.0 & 7.1 \\
Qwen3.5-9B & - & 85.8 & 85.4 & {\bfseries 86.2} & -0.7 & & {\bfseries 91.9} & 79.7 & +12.3 & & {\bfseries 7.7} & 1.6 & 4.9 \\
Qwen3.5-27B & - & 96.1 & 95.8 & {\bfseries 96.3} & -0.6 & & {\bfseries 96.8} & 95.4 & +1.4 & & {\bfseries 2.8} & 0.7 & 0.4 \\
\hline
\tabH Qwen3.5-0.8B & \checkmark & 33.3 & 32.3 & {\bfseries 34.2} & -1.9 & & 30.6 & {\bfseries 36.0} & -5.4 & & {\bfseries 25.8} & 20.1 & 20.9 \\
Qwen3.5-2B & \checkmark & 80.3 & 78.2 & {\bfseries 82.3} & -4.1 & & {\bfseries 81.3} & 79.3 & +2.0 & & {\bfseries 13.1} & 4.3 & 2.3 \\
Qwen3.5-4B & \checkmark & 95.5 & 94.7 & {\bfseries 96.3} & -1.6 & & {\bfseries 96.2} & 94.8 & +1.4 & & {\bfseries 3.8} & 0.4 & 0.2 \\
Qwen3.5-9B & \checkmark & 97.3 & 96.7 & {\bfseries 97.9} & -1.2 & & {\bfseries 98.2} & 96.5 & +1.7 & & {\bfseries 2.4} & 0.2 & 0.1 \\
Qwen3.5-27B & \checkmark & 98.6 & 98.1 & {\bfseries 99.1} & -1.0 & & {\bfseries 99.1} & 98.1 & +1.0 & & {\bfseries 1.0} & 0.3 & 0.1 \\
\bhline
\end{tabular}
\caption{
Breakdown of English FrameBench performance by sentence-pair type, position of the correct sentence, and error type.
$\mathcal{R}$ denotes reasoning mode, and Acc. denotes overall accuracy.
Base and Ext. report accuracy on base and extended sentence pairs, while 1st and 2nd report accuracy when the correct sentence appears first or second.
$\Delta_{B-E}$ and $\Delta_{1-2}$ indicate performance differences. 
Both denotes selecting both sentences, Neither denotes selecting neither sentence, and Opposite denotes selecting the incorrect sentence.
Acc. and the three error-type rates sum to 100\% up to rounding.
}
\label{tab:discussion}
\end{table*}
\begin{table}[h]
\centering
\fontsize{8pt}{9pt}\selectfont
\begin{tabular}{lcccc}
\bhline
\tabH Model & $\mathcal{R}$ & $\mathcal{M}$ & \makecell[c]{\tabH Frame\\Bench} & \makecell[c]{\tabH MMLU-\\Pro} \\ \bhline
\tabH Qwen3-1.7B   & \checkmark &            & \textbf{77.8} & 35.6$^{c}$ \\
Qwen3-VL-2B  & \checkmark & \checkmark & 67.7 & \textbf{62.3}$^{a}$ \\ \hline
\tabH Qwen3-4B     & \checkmark &            & 90.9 & 52.2$^{c}$ \\
Qwen3-VL-4B  & \checkmark & \checkmark & \textbf{92.3} & \textbf{73.6}$^{a}$ \\ \hline
\tabH Qwen3-8B     & \checkmark &            & 92.5 & 72.1$^{c}$ \\
Qwen3-VL-8B  & \checkmark & \checkmark & \textbf{96.3} & \textbf{77.3}$^{a}$ \\ \hline
\tabH Qwen3-32B    & \checkmark &            & \textbf{96.9} & 66.8$^{c}$ \\ 
Qwen3-VL-32B & \checkmark & \checkmark & 95.7 & \textbf{82.1}$^{a}$ \\ \hline
\tabH Qwen3-1.7B   &            &            & 53.5 & 37.3$^{f}$ \\
Qwen3-VL-2B  &            & \checkmark & \textbf{71.4} & \textbf{49.0}$^{a}$ \\ \hline
\tabH Qwen3-4B     &            &            & 84.6 & 57.9$^{f}$ \\ 
Qwen3-VL-4B  &            & \checkmark & \textbf{89.8} & \textbf{67.1}$^{a}$ \\ \hline
\tabH Qwen3-8B     &            &            & 87.1 & 59.9$^{f}$ \\
Qwen3-VL-8B  &            & \checkmark & \textbf{94.3} & \textbf{71.6}$^{a}$ \\ \hline
\tabH Qwen3-32B    &            &            & 92.5 & 72.7$^{f}$ \\
Qwen3-VL-32B &            & \checkmark & \textbf{95.7} & \textbf{78.6}$^{a}$ \\ \hline
\tabH Phi-3.5-mini     &            &            & 59.7 & \textbf{34.5}$^{f}$ \\
Phi-3.5-vision   &            & \checkmark & \textbf{72.1} & 32.2$^{f}$ \\ \hline
\tabH Phi-4-mini       &            &            & 78.5 & \textbf{43.3}$^{f}$ \\
Phi-4-multimodal &            & \checkmark & \textbf{79.8} & 39.1$^{f}$ \\ \bhline
\end{tabular}
\caption{
Performance comparison of matched text-only and multimodal models.
$\mathcal{R}$ denotes reasoning mode, and $\mathcal{M}$ denotes multimodality.
For MMLU-Pro scores, superscripts indicate the source of each score: 
$^a$ model-provider reports, 
$^c$ \href{https://artificialanalysis.ai/}{Artificial Analysis}, 
and $^f$ our own evaluations.}
\label{tab:vl_comperision}
\end{table}

\section{Analysis}
\after{We analyze model behavior on the English subset of FrameBench to better understand error patterns and remaining challenges.}

\begin{table*}[t]
\centering
\fontsize{8pt}{9pt}\selectfont
\setlength{\tabcolsep}{1.2pt}
\renewcommand{\arraystretch}{1.15}
\begin{tabular}{>{\raggedright\arraybackslash}m{0.66\textwidth}@{\hspace{1.5em}}*{5}{>{\centering\arraybackslash}m{0.45cm}}@{\hspace{1.5em}}*{3}{>{\centering\arraybackslash}m{0.45cm}}@{\hspace{0.5em}}}
\bhline
\multirow{2}{0.66\textwidth}{\centering\tabH Question and Sentence Pair} & \multicolumn{5}{c}{\tabH Qwen3.5} & \multicolumn{3}{c}{\tabH Gemma~4} \\
& 0.8B & 2B & 4B & 9B & 27B & E2B & E4B & 31B \\
\bhline
\textbf{Q1.} Select all sentences where the speaker gets hurt. \par
\textbf{$S_{\text{A}}$.} Rounding the corner too fast, I \uline{grazed} the wall with my knee.
{\footnotesize(\texttt{[Impact]})} \par
\textbf{$S_{\text{B}}$.} \textbf{Rounding the corner too fast, I \uline{grazed} my knee on the wall.
{\footnotesize(\texttt{[Body\_injury]})}}
& {\scriptsize--} & {\scriptsize--} & {\scriptsize--} & {\scriptsize--} & {\scriptsize--} & {\scriptsize--} & {\scriptsize$\checkmark$} & {\scriptsize--} \\ \hline
\textbf{Q2.} Select all sentences where someone carries out a ceremonial rite. \par
\textbf{$S_{\text{A}}$.} They \uline{christened} their new boat the Sea Breeze at the pier.
{\footnotesize(\texttt{[Name\_conferral]})} \par
\textbf{$S_{\text{B}}$.} \textbf{They \uline{christened} their new boat with a bottle of champagne at the pier.
{\footnotesize(\texttt{[Rite]})}}
& {\scriptsize$\checkmark$} & {\scriptsize--} & {\scriptsize--} & {\scriptsize--} & {\scriptsize--} & {\scriptsize--} & {\scriptsize--} & {\scriptsize--} \\
\hline
\textbf{Q3.} Select all sentences where someone is trying to follow the person. \par
\textbf{$S_{\text{A}}$.} \textbf{I \uline{lost} the guards in the crowd.
{\footnotesize(\texttt{[Losing\_track\_of\_perceiver]})}} \par
\textbf{$S_{\text{B}}$.} I \uline{lost} my wallet in the crowd.
{\footnotesize(\texttt{[Losing]})}
& {\scriptsize--} & {\scriptsize--} & {\scriptsize$\checkmark$} & {\scriptsize$\checkmark$} & {\scriptsize$\checkmark$} & {\scriptsize--} & {\scriptsize$\checkmark$} & {\scriptsize$\checkmark$} \\
\hline
\textbf{Q4.} Select all sentences where a person takes on a character in a play. \par
\textbf{$S_{\text{A}}$.} \textbf{He and Maya \uline{played} Romeo and Juliet at the festival.
{\footnotesize(\texttt{[Performers\_and\_roles]})}} \par
\textbf{$S_{\text{B}}$.} He and Maya \uline{played} tennis at the festival.
{\footnotesize(\texttt{[Competition]})}
& {\scriptsize$\checkmark$} & {\scriptsize$\checkmark$} & {\scriptsize$\checkmark$} & {\scriptsize$\checkmark$} & {\scriptsize$\checkmark$} & {\scriptsize$\checkmark$} & {\scriptsize$\checkmark$} & {\scriptsize$\checkmark$} \\
\bhline
\end{tabular}
\caption{
Examples from FrameBench.
Bold text marks the correct answer, underlining marks the frame-evoking word, and parenthetical tags show the corresponding frame.
Each model column reports whether the model answered the item correctly in the reasoning setting,
where `\checkmark' denotes a correct prediction and `--' denotes an incorrect prediction.
}
\label{tab:examples}
\end{table*}

\subsection{Error and Bias Analysis}
\label{subsec:discussion}
\after{We further analyze model behavior on FrameBench along three dimensions: sentence-pair type, sentence position, and error type.
Japanese results for the same analyses are provided in Appendix~\ref{app:ja_analysis}; they show similar tendencies except for a weaker sentence-pair type effect.
}

\paragraph{Difficulty by Sentence Pair Type}
\after{Since base pairs are designed to be lexically similar, they are expected to be harder for models than extended pairs, which are not constrained to be surface-similar.
Columns under \textbf{(i) Sent. Pair Type} in Table~\ref{tab:discussion} show the accuracy for each pair type and the performance gap. 
All models achieved higher accuracy on extended pairs than on base pairs, as indicated by the negative values of $\Delta_{B-E}$. 
The human score gap shows the same tendency, with extended pairs outperforming base pairs by 2.57 points.
These results indicate that lexical similarity makes base pairs harder for LLMs.
In the Japanese subset, however, this pair-type effect is weaker: the human score gap decreases to 0.93 points, and the LLM results show a less consistent advantage for extended pairs.}

\paragraph{Analysis of Sentence Position Bias}
\after{We also investigated order sensitivity by comparing performance when the correct sentence appears in the first versus second position. 
Columns under \textbf{(ii) Correct Sent. Position} in Table~\ref{tab:discussion} report accuracy for these two positions and the corresponding performance gap.
Since each sentence pair is evaluated in both original and swapped orders, the same items appear in both position conditions, controlling for item difficulty. 
Most models show a noticeable absolute difference between the two positions, indicating that sentence order can affect model predictions. 
This supports the main evaluation protocol described in Section~\ref{sec:experiment}, where each pair is evaluated in both possible orders to mitigate order sensitivity.}

\paragraph{Error Tendencies in Incorrect Choices}
\after{
Columns under \textbf{(iii) Error Breakdown} in Table~\ref{tab:discussion} show the distribution of model errors. 
For models scoring above 85\%, ``Opposite'' errors become rare.
For high-performing models, errors therefore tend to reflect over-selection or under-selection rather than outright selection of the opposite frame, suggesting that the remaining difficulty lies in borderline frame-semantic distinctions.
}

\subsection{Impact of Multimodality}
\label{subsec:multimodality}

\after{Frame semantics is a linguistic framework, but the situations it describes often involve visually grounded event knowledge.}
\after{This motivates examining whether multimodal training helps models interpret frame-semantic distinctions.}
\after{Table~\ref{tab:vl_comperision} reports matched comparisons between multimodal models and the language models on which they are based.
}
\after{Overall, multimodal models outperform the corresponding language models in most matched comparisons on FrameBench.}
\after{However, these gains should be interpreted cautiously: many multimodal models also achieve higher MMLU-Pro scores, while the Phi models show FrameBench improvements despite lower MMLU-Pro scores.}
\after{Thus, visually grounded training may help frame-semantic interpretation, but its effect cannot be isolated from other training differences.}


\subsection{Case Study}
\after{
We examine examples selected to span different levels of model accuracy.
Because we consider only items for which at least two of the three human annotators selected the correct answer, these examples are intended to be answerable by humans rather than inherently ambiguous.
Table~\ref{tab:examples} reports per-model correctness for reasoning-enabled Qwen3.5 and Gemma~4 models using the first prompt template in Table~\ref{tab:prompts_en}.
The scores are shown across model sizes, allowing us to examine whether each example is solved consistently, requires larger models, or remains difficult even for stronger models.
}

\after{In Questions 1 and 2 of Table~\ref{tab:examples}, even the largest Qwen3.5 and Gemma~4 models fail, incorrectly selecting both sentences.
These examples require distinguishing subtle frame-semantic differences despite strong lexical or contextual overlap.
In Question 1, a small change in argument structure shifts the interpretation from contact with an object to injury to a body part.
In Question 2, \textit{christen} carries a ceremonial nuance in both sentences, but only Sentence B explicitly evokes the \texttt{[Rite]} frame through the use of a bottle of champagne.
These results suggest that even strong models can struggle when the relevant frame distinction depends on fine-grained semantic cues.}

\after{Question 3 is easier, but still challenging for smaller models.
Both sentences use \textit{lost} in closely related senses, but only Sentence A describes a situation in which someone trying to follow the speaker loses track of them.
The score pattern suggests that distinguishing related senses within a close semantic domain requires a moderate level of semantic sensitivity.}

\after{By contrast, Question 4 is solved by all models in the table.
Here, the two meanings of \textit{play} belong to clearly different semantic domains, and the distinction can be identified from salient contextual cues such as \textit{Romeo and Juliet} and \textit{tennis}.}



\section{Conclusion}

In this study, we introduced FrameBench, a FrameNet-grounded multiple-choice benchmark for evaluating whether LLMs can distinguish the frames evoked by the same verb across contexts.
We constructed English and Japanese versions of FrameBench using FrameNet resources, a generation-and-verification pipeline, and native-speaker judgments.
Together, these datasets provide a new resource for evaluating frame-semantic interpretation in LLMs across two typologically distant languages.

Experiments across a wide range of LLMs showed that several frontier models reached or exceeded the human reference score, while performance varied substantially with model scale, reasoning mode, and language.
Our analyses further characterized model behavior in terms of sentence-pair type, sentence position, and error type, and examined whether multimodal training may support frame-semantic interpretation.
\section*{Limitations}
First, FrameBench focuses on a specific aspect of semantic competence, namely frame-semantic discrimination for context-dependent verb interpretations in a four-choice setting.
It does not directly evaluate broader language understanding, open-ended generation, or explicit frame prediction.
Second, the human reference scores reported in our experiments are not based on an independent evaluation.
They are calculated from the judgments of the same annotators whose responses were used to validate and filter FrameBench items.
Because our experiments use only items that were answered correctly by at least two annotators during validation, the reported human scores may overestimate performance relative to an evaluation conducted with an independent group of annotators.

\section*{Ethical considerations}
Our dataset construction involved human annotation in both English and Japanese. 
English annotations were conducted by three expert native English annotators recruited through an annotation vendor and compensated according to the vendor’s standard rates. 
Japanese evaluation was conducted by three native Japanese-speaking university students, who were compensated at the university-defined hourly rate for research assistants. 
Manual revision of Japanese generations was performed by one of the authors as part of the research process.
All annotators were informed in advance that their judgments would be used for research purposes and reported in a paper.

\section*{Acknowledgments}
\cameraready{We would like to express our gratitude to Dr. Kyoko Ohara of Keio University for providing the Japanese FrameNet data used in this study. 
This research was supported by JST FOREST Program JPMJFR216N and JST SPRING Program JPMJSP2125.}
\bibliography{custom}

\appendix

\section{Details of Human Annotation and Manual Revision}
\subsection{Human Annotation Protocol}
We manually evaluated correctness and acceptability for the finalized items in both English and Japanese FrameBench.
For correctness, annotators answered the same four-choice question format used in model evaluation.
For acceptability, Japanese items were judged with a binary naturalness label, whereas English items were judged with a three-way scale (\textit{Unacceptable}, \textit{Acceptable}, \textit{Natural}) to reduce rater drift.

\subsection{Acceptability scale and binarization}
To facilitate cross-lingual analysis, we convert acceptability judgments into a binary label.
For English, we map \textit{Natural} to the positive label and treat \textit{Acceptable} and \textit{Unacceptable} as negative.
This mapping yields a more informative label distribution than treating only \textit{Unacceptable} as negative, since \textit{Unacceptable} is extremely rare in our data, and it aligns with our goal of using a stricter notion of linguistic naturalness.
We report the three-way label distribution for English in Table~\ref{tab:acceptability_3way_dist_en}.

\begin{table}[h]
\centering
\small
\setlength{\tabcolsep}{4pt}
\begin{tabular}{lccc}
\bhline
 & \tabH \textbf{Natural} & \textbf{Acceptable} & \textbf{Unacceptable} \\
\hline
\tabH Ann1 & 83.8 & 14.9 & 1.3 \\
Ann2 & 91.4 & 8.3  & 0.2 \\
Ann3 & 82.6 & 16.8 & 0.6 \\
\bhline
\end{tabular}
\caption{Distribution of 3-level acceptability labels for English. Values are percentages.}
\label{tab:acceptability_3way_dist_en}
\end{table}
\begin{table}[h]
\centering
\small
\setlength{\tabcolsep}{3pt}
\begin{tabular}{lccc cc}
\bhline
 & \multicolumn{3}{c}{\tabH Positive label (\%)} & Agreement& Fleiss' \\
& Ann1 & Ann2 & Ann3 & (\%)  &$\kappa$ \\
\hline
\multicolumn{6}{l}{\tabH  \textbf{English}}\\
\hline
\multicolumn{6}{l}{\tabH  \textbf{Correctness}}\\
Base     & 88.8 & 92.5 & 93.6 & 82.5 & 0.2406 \\
Extended & 93.8 & 96.0 & 96.0 & 89.3 & 0.2054 \\
\hdashline
\multicolumn{6}{l}{\tabH  \textbf{Acceptability}}\\
Base     & 82.9 & 91.8 & 81.0 & 67.6 & 0.1417 \\
Extended & 84.5 & 91.9 & 83.2 & 68.8 & 0.1070 \\
\hline
\multicolumn{6}{l}{\tabH  \textbf{Japanese}}\\
\hline
\multicolumn{6}{l}{\tabH  \textbf{Correctness}}\\
Base     & 86.2 & 87.3 & 96.2 & 77.2 & 0.1670 \\
Extended & 91.1 & 91.6 & 96.4 & 84.5 & 0.2054 \\
\hdashline
\multicolumn{6}{l}{\tabH \textbf{Acceptability}}\\
Base     & 96.0 & 99.1 & 97.1 & 92.9 & 0.0687 \\
Extended & 98.0 & 99.5 & 97.1 & 94.9 & 0.0494 \\
\bhline
\end{tabular}
\caption{Inter-annotator agreement for binary judgments.
Agreement is the percentage of pairs where all three annotators made the same judgment.}
\label{tab:iaa_binary_by_pair}
\end{table}

\subsection{Inter-annotator agreement}
\label{app:annotator-agreement}
Table~\ref{tab:iaa_binary_by_pair} reports positive-label rates, Fleiss' $\kappa$, and the 3/3 agreement rate for the binary judgments.
Because $\kappa$ is sensitive to skewed label distributions, especially for acceptability, we report these additional statistics to aid interpretation.
In English, acceptability contains more borderline cases than correctness, which can lower agreement.
For the main experiments, we therefore use a high-quality subset defined by a strict human validation criterion.
Under this criterion, an entry is included only if at least two annotators judged it correct and at least two judged it acceptable.

\section{Experimental Details}
\label{app:experimental_details}

\subsection{Evaluation Prompts}

Table \ref{tab:prompts_en} and Table \ref{tab:prompts_ja} present the prompts used for English and Japanese evaluation. 
The variables \{question\}, \{verb\}, \{sentence\_a\}, \{sentence\_b\}, and \{choices\_text\} serve as placeholders for the question text, the target verb, the target sentence pair, and the options, respectively. 
To eliminate the influence of specific choice number output probabilities, the mapping between options and numbers was randomized for each question.

\begin{table}[h]
\centering
\small
\begin{tabularx}{\linewidth}{>{\raggedright\arraybackslash}X}
\bhline
\tabH \textbf{Prompt templates} \\
\hline
\tabH
\{question\}\par
Sentence A: \{sentence\_a\}\par
Sentence B: \{sentence\_b\}\par
Choices: \{choices\_text\}\par
Instruction: Please make your decision by focusing on the verb ``\{verb\}''.\par
Respond using only the choice number: ``1'', ``2'', ``3'', or ``4''. \\
\hline
\tabH
Please make your decision by focusing on the verb ``\{verb\}''.\par
\{question\}\par
Sentence A: \{sentence\_a\}\par
Sentence B: \{sentence\_b\}\par
Choices: \{choices\_text\}\par
Respond using only the choice number: ``1'', ``2'', ``3'', or ``4''. \\
\hline
\tabH
\{question\}\par
Sentence A: \{sentence\_a\}\par
Sentence B: \{sentence\_b\}\par
Choices: \{choices\_text\}\par
Respond using only the choice number: ``1'', ``2'', ``3'', or ``4''. \\
\hline
\tabH
Your task is to compare the two sentences through the usage of ``\{verb\}''.\par
\{question\}\par
Sentence A: \{sentence\_a\}\par
Sentence B: \{sentence\_b\}\par
Available choices:\par
\{choices\_text\}\par
Output strictly one number (1--4). No explanation. \\
\hline
\tabH
Problem: \{question\}\par
[Sentence A]\par
\{sentence\_a\}\par
[Sentence B]\par
\{sentence\_b\}\par
Choices:\par
\{choices\_text\}\par
Final answer format: just the option number. \\
\bhline
\end{tabularx}
\caption{Prompt templates used for English evaluation.}
\label{tab:prompts_en}
\end{table}
\label{app:prompt}

\begin{table}[h!]
\centering
\small
\begin{tabularx}{\linewidth}{>{\raggedright\arraybackslash}X}
\bhline
\tabH \textbf{Prompt templates} \\
\hline
\tabH 
\{question\}\par
文A: \{sentence\_a\}\par
文B: \{sentence\_b\}\par
選択肢: \{choices\_text\}\par
回答する際は、文の最後の動詞に注目して判断してください。\par
\hspace{1em}\textit{(When answering, focus on the sentence-final verb in each sentence.)}\par
回答は選択肢の番号「1」「2」「3」「4」のいずれかで答えてください。\par
\hspace{1em}\textit{(Respond using only the option number 1, 2, 3, or 4.)}\\
\hline
\tabH
\{question\}\par
文A: \{sentence\_a\}\par
文B: \{sentence\_b\}\par
選択肢: \{choices\_text\}\par
選択肢の番号「1」「2」「3」「4」のいずれかで答えてください。\par
\hspace{1em}\textit{(Answer with the option number 1, 2, 3, or 4.)}\\
\hline
\tabH
それぞれの文の述語動詞に注意して、\{question\}\par
\hspace{1em}\textit{(Pay attention to the predicate verb in each sentence: \{question\})}\par
文A: \{sentence\_a\}\par
文B: \{sentence\_b\}\par
選択肢: \{choices\_text\}\par
選択肢の番号「1」「2」「3」「4」のいずれかで答えてください。\par
\hspace{1em}\textit{(Answer with the option number 1, 2, 3, or 4.)}\\
\hline
\tabH
それぞれの文の述語動詞に注意して回答してください。\par
\hspace{1em}\textit{(Pay attention to the predicate verb in each sentence.)}\par
\{question\}\par
文A: \{sentence\_a\}\par
文B: \{sentence\_b\}\par
選択肢: \{choices\_text\}\par
選択肢の番号「1」「2」「3」「4」のいずれかで答えてください。\par
\hspace{1em}\textit{(Answer with the option number 1, 2, 3, or 4.)}\\
\hline
\tabH
選択肢の番号で回答してください。\par
\hspace{1em}\textit{(Please answer using the number of the correct option.)}\par
\{question\}\par
文A: \{sentence\_a\}\par
文B: \{sentence\_b\}\par
選択肢: \{choices\_text\} \\
\bhline

\end{tabularx}
\caption{Prompt templates used for Japanese evaluation.}
\label{tab:prompts_ja}
\end{table}

\subsection{\after{Evaluated Models List}}
\label{app:model_list}

\textbf{Closed-source Models}
\begin{itemize}[nosep]
    \item \textbf{GPT-5 Series}~\cite{GPT5}: \texttt{gpt-5-2025-08-07}, \texttt{gpt-5-nano-2025-08-07}
    \item \textbf{Gemini 3.1 Series}: \href{https://blog.google/innovation-and-ai/models-and-research/gemini-models/gemini-3-1-pro/}{\texttt{gemini-3.1-pro}}, \href{https://deepmind.google/models/model-cards/gemini-3-1-flash-lite/}{\texttt{gemini-3.1-flash-lite}}
\end{itemize}

\textbf{Open-weight Models}
\begin{itemize}[nosep]

    \item \textbf{Gemma 4 Series}~\cite{gemma4}: 
    \href{https://huggingface.co/google/gemma-4-31B-it}{Gemma-4-31B-it}, 
    \href{https://huggingface.co/google/gemma-4-E4B-it}{E4B-it}, 
    \href{https://huggingface.co/google/gemma-4-E2B-it}{E2B-it}

    \item \textbf{Qwen 3.5 Series}~\cite{qwen3.5}: 
    \href{https://huggingface.co/Qwen/Qwen3.5-27B}{Qwen3.5-27B}, 
    \href{https://huggingface.co/Qwen/Qwen3.5-9B}{9B}, 
    \href{https://huggingface.co/Qwen/Qwen3.5-4B}{4B}, 
    \href{https://huggingface.co/Qwen/Qwen3.5-2B}{2B}, 
    \href{https://huggingface.co/Qwen/Qwen3.5-0.8B}{0.8B}
    
    \item \textbf{Qwen 3 Series}~\cite{Qwen3}: 
    \href{https://huggingface.co/Qwen/Qwen3-1.7B}{Qwen3-1.7B}, 
    \href{https://huggingface.co/Qwen/Qwen3-4B}{Qwen3-4B}, 
    \href{https://huggingface.co/Qwen/Qwen3-8B}{Qwen3-8B}, 
    \href{https://huggingface.co/Qwen/Qwen3-32B}{Qwen3-32B}

    \item \textbf{Phi Series}: 
    \href{https://huggingface.co/microsoft/Phi-3.5-mini-instruct}{Phi-3.5-mini-instruct}~\cite{phi3}, 
    \href{https://huggingface.co/microsoft/Phi-4-mini-instruct}{Phi-4-mini-instruct}~\cite{phi4}
\end{itemize}

\textbf{Japanese-oriented Models}
\begin{itemize}[nosep]
    \item \textbf{LLM-jp-4 Series}~\cite{llm-jp}: 
    \href{https://huggingface.co/llm-jp/llm-jp-4-8b-thinking}{llm-jp-4-8b-thinking}, 
    \href{https://huggingface.co/llm-jp/llm-jp-4-8b-instruct}{8b-instruct}

    \item \textbf{Swallow Series}~\cite{qwen3-swallow}: 
    \href{https://huggingface.co/tokyotech-llm/Qwen3-Swallow-8B-RL-v0.2}{Qwen3-Swallow-8B-RL}, 
    \href{https://huggingface.co/tokyotech-llm/Qwen3-Swallow-32B-RL-v0.2}{32B-RL}\\
    Qwen3-Swallow models are developed through continual pre-training, SFT, and RL based on Qwen3-Base to enhance their Japanese language capabilities.
\end{itemize}

\textbf{Multimodal Models}
\begin{itemize}[nosep]
\item \textbf{Qwen3-VL Series}~\cite{Qwen3-VL}: 
    \href{https://huggingface.co/Qwen/Qwen3-VL-2B-Thinking}{Qwen3-VL-2B-Thinking}, 
    \href{https://huggingface.co/Qwen/Qwen3-VL-4B-Thinking}{4B-Thinking}, 
    \href{https://huggingface.co/Qwen/Qwen3-VL-8B-Thinking}{8B-Thinking}, 
    \href{https://huggingface.co/Qwen/Qwen3-VL-32B-Thinking}{32B-Thinking}, 
    \href{https://huggingface.co/Qwen/Qwen3-VL-2B-Instruct}{2B-Instruct}, 
    \href{https://huggingface.co/Qwen/Qwen3-VL-4B-Instruct}{4B-Instruct}, 
    \href{https://huggingface.co/Qwen/Qwen3-VL-8B-Instruct}{8B-Instruct}, 
    \href{https://huggingface.co/Qwen/Qwen3-VL-32B-Instruct}{32B-Instruct}\\
Qwen3-VL models are built on Qwen3-family language backbones and incorporate vision-specific components to support text and image inputs.
    \item \textbf{Phi Series}: 
    \href{https://huggingface.co/microsoft/Phi-3.5-vision-instruct}{Phi-3.5-vision-instruct}~\cite{phi3}, 
    \href{https://huggingface.co/microsoft/Phi-4-multimodal-instruct}{Phi-4-multimodal-instruct}~\cite{phi4}\\
Both Phi-3.5-vision-instruct and Phi-4-multimodal-instruct are multimodal Phi-series models that support text and image inputs, using language backbones derived from Phi-3.5-mini and Phi-4-Mini-Instruct, respectively.
\end{itemize}

\subsection{Decoding Procedure}
\label{app:answer_extraction}
For open-weight models evaluated in the non-reasoning setting, generation is restricted to the tokens corresponding to the answer choices. 
\after{In the reasoning setting, models with reasoning capabilities generate tokens freely until they output a specific tag marking the end of the reasoning phase. After this tag is generated, decoding is restricted in the same way as in the non-reasoning setting, allowing the model to generate only the tokens corresponding to the valid answer choices.
For models with a native reasoning process, such as the GPT-5 series, we utilize JSON-based structured output to facilitate reliable answer extraction.}

\subsection{Details of the Comparison Benchmarks}
\label{app:benchmarks}
We describe the LLM benchmarks used for comparison in each experiment.
Unless otherwise noted, we adopt the evaluation protocol in \texttt{swallow-evaluation-instruct}\footnote{\cameraready{Version: v202510}}~\cite{swallow-evaluation-instruct}.

\textbf{English Benchmarks}
\label{app:english_benchmarks}

\begin{itemize}[nosep]
    \item \textbf{MMLU-Pro~\cite{mmlu-pro}:}
A harder variant of MMLU that evaluates broad domain knowledge via multiple-choice questions.
It spans 14 subject areas, including humanities, social sciences, and STEM.

\item \textbf{GPQA-Diamond~\cite{gpqa}:}
A four-choice, graduate-level science QA benchmark written by domain experts. 
It covers biology, physics, and chemistry and is designed to be difficult to solve via web search.

\item \textbf{ Humanity's Last Exam~\cite{hle}:}
\after{A cross-disciplinary benchmark comprising 2,500 questions across dozens of subjects, including mathematics, the humanities, and the natural sciences.
It features a mix of multiple-choice and short-answer questions designed to evaluate expert-level reasoning across broad academic domains.}
\end{itemize}




\textbf{Japanese Benchmarks}
\label{app:japanese_benchmarks}
\begin{itemize}[nosep]
\item \textbf{JamC-QA~\cite{jamcqa}:}
A high-difficulty four-choice QA benchmark specialized for knowledge of Japan-specific culture and customs.
Questions cover eight categories including culture, customs, local environment, geography, administration, law, medicine, and related topics, requiring niche and diverse knowledge.
\item \textbf{MMLU-ProX~\cite{mmluprox}:}
A multilingual extension of MMLU-Pro covering 29 languages. We use its Japanese subset.
\end{itemize}







\subsection{Details of the Frame Identification Evaluation}
\label{app:frameid}

\after{
This section describes the frame identification evaluation.
Unlike FrameBench, this evaluation directly provides models with candidate frame names and their definitions from the corresponding frame-semantic resources: FrameNet for English and Japanese FrameNet for Japanese.
It therefore tests whether models can identify the frame evoked by a target word when the relevant candidate frames are explicitly given.
}

\after{
Each test instance consists of a sentence containing a target frame-evoking verb and a set of answer options.
The answer options include two candidate frames that the same verb can evoke, as well as a negative option, ``Neither frame is evoked.''
The order of the answer options is randomized.}

\after{
We construct this evaluation from the same verb--frame pairs used in FrameBench.
When a verb appears in multiple FrameBench pairs because it can evoke three or more frames, we randomly select one pair for that verb to avoid redundancy.
For each selected verb--frame pair, we sample one sentence for each of the two candidate frames from the corresponding frame-semantic resource.
Thus, each pair yields two independent frame identification instances, one for each candidate frame.}

\after{
We evaluate 200 verb--frame pairs in a 3-shot setting.
Because each pair contributes two independently evaluated target sentences, the resulting evaluation set contains 400 test instances in total.}

\section{Effect of the LLM Used for Dataset Construction}
\label{app:generation_model}

\cameraready{
To assess the effect of the LLM used in dataset construction, we generated an additional 100 English candidate FrameBench entries using Gemini 3.1 Pro in place of GPT-5 for Steps 1 and 2.
The verb--frame pairs used for these entries were selected from those used in the original GPT-5-based construction.
To limit additional annotation costs, we did not conduct the human validation and filtering in Step 3 for these additional entries.
We evaluated them using the same model evaluation protocol as in the main experiment.
}

\begin{table}[t]
\centering
\setlength{\tabcolsep}{5pt}
\small
\begin{tabular}{lcc}
\bhline
\tabH \multirow{2}{*}{Model} 
 & \multirow{2}{*}{$\mathcal{R}$} & FrameBench \\ 
  & & (Gemini-generated)  \\ \bhline
\tabH GPT-5 nano & \checkmark & 92.7 \\
GPT-5 & \checkmark & 97.9 \\
Gemini 3.1 Flash-Lite & \checkmark & 98.8 \\
Gemini 3.1 Pro & \checkmark & 100.0 \\
\hline
\tabH Gemma 4 E2B & \checkmark & 88.6 \\
Gemma 4 E4B & \checkmark & 95.5 \\
Gemma 4 31B & \checkmark & 98.9 \\
Gemma 4 E2B & -- & 66.2 \\
Gemma 4 E4B & -- & 85.5 \\
Gemma 4 31B & -- & 97.9 \\
\hline
\tabH Qwen3.5-0.8B & \checkmark & 33.8 \\
Qwen3.5-2B & \checkmark & 80.8 \\
Qwen3.5-4B & \checkmark & 96.6 \\
Qwen3.5-9B & \checkmark & 96.9 \\
Qwen3.5-27B & \checkmark & 99.1 \\
Qwen3.5-0.8B & -- & 47.6 \\
Qwen3.5-2B & -- & 47.6 \\
Qwen3.5-4B & -- & 82.4 \\
Qwen3.5-9B & -- & 86.9 \\
Qwen3.5-27B & -- & 97.1 \\
\bhline
\end{tabular}
\caption{English FrameBench results on 100 entries generated using Gemini 3.1 Pro instead of GPT-5 in Steps 1 and 2. $\mathcal{R}$ denotes reasoning mode.}
\label{tab:gemini_generated}
\end{table}

\cameraready{
Table~\ref{tab:gemini_generated} reports the results.
We compared the results on entries constructed using Gemini 3.1 Pro with the original FrameBench results reported in Table~\ref{tab:result_merged}.
Performance tended to be slightly higher for models from the same family as the LLM used to construct the benchmark.
However, the overall performance pattern was similar across the two sets, with a Spearman rank correlation of $\rho=0.986$.}
\begin{table*}[t!]
\centering
\fontsize{8pt}{9pt}\selectfont
\begin{tabular}{
  l c
  S[table-format=2.1]
  S[table-format=2.1] S[table-format=2.1] S[table-format=-1.1]
  p{0.05cm}
  S[table-format=2.1] S[table-format=2.1] S[table-format=+2.1]
  p{0.05cm}
  S[table-format=2.1] S[table-format=2.1] S[table-format=2.1]
}
\bhline
&&& \multicolumn{3}{c}{\tabH (i) Sent. Pair Type} & & \multicolumn{3}{c}{\tabH (ii) Correct Sent. Position} & & \multicolumn{3}{c}{\tabH (iii) Error Breakdown} \\ 
\cline{4-6} \cline{8-10} \cline{12-14}
\tabH Model & $\mathcal{R}$ & \multicolumn{1}{c}{\tabH Acc.} & \multicolumn{1}{c}{\tabH Base} & \multicolumn{1}{c}{\tabH Ext.} & \multicolumn{1}{c}{\tabH $\Delta_{B-E}$} & & \multicolumn{1}{c}{\tabH 1st} & \multicolumn{1}{c}{\tabH 2nd} & \multicolumn{1}{c}{\tabH $\Delta_{1-2}$} & & \multicolumn{1}{c}{\tabH Both} & \multicolumn{1}{c}{\tabH Neither} & \multicolumn{1}{c}{\tabH Opposite} \\ \bhline
\tabH Gemma~4~E2B & - & 41.5 & 38.8 & {\bfseries 44.2} & -5.5 &  & {\bfseries 53.4} & 29.6 & +23.8 &  & 20.1 & {\bfseries 22.1} & 16.4 \\
Gemma~4~E4B & - & 84.6 & 84.3 & {\bfseries 84.8} & -0.4 &  & 84.0 & {\bfseries 85.0} & -1.0 &  & {\bfseries 5.9} & 4.0 & 5.6 \\
Gemma~4~31B & - & 97.6 & {\bfseries 97.7} & 97.5 & +0.2 &  & {\bfseries 98.5} & 96.8 & +1.6 &  & 0.7 & {\bfseries 1.2} & 0.4 \\
\hline
\tabH Gemma~4~E2B & \checkmark & 82.3 & 78.6 & {\bfseries 86.0} & -7.5 &  & 80.7 & {\bfseries 84.0} & -3.3 &  & {\bfseries 8.5} & 8.3 & 0.8 \\
Gemma~4~E4B & \checkmark & 92.2 & 90.9 & {\bfseries 93.5} & -2.6 &  & 92.0 & {\bfseries 92.5} & -0.5 &  & {\bfseries 4.0} & 3.1 & 0.7 \\
Gemma~4~31B & \checkmark & 99.1 & {\bfseries 99.2} & 99.0 & +0.2 &  & {\bfseries 99.5} & 98.6 & +0.9 &  & {\bfseries 0.4} & 0.3 & 0.1 \\
\hline
\tabH Qwen3.5-0.8B & - & 25.4 & {\bfseries 25.5} & 25.4 & +0.1 &  & {\bfseries 27.8} & 23.1 & +4.8 &  & {\bfseries 28.8} & 20.4 & 25.4 \\
Qwen3.5-2B & - & 30.0 & 28.8 & {\bfseries 31.3} & -2.5 &  & {\bfseries 34.5} & 25.5 & +8.9 &  & 21.9 & {\bfseries 28.0} & 20.1 \\
Qwen3.5-4B & - & 56.7 & {\bfseries 57.6} & 55.7 & +1.9 &  & {\bfseries 66.5} & 46.8 & +19.6 &  & 12.1 & {\bfseries 19.0} & 12.3 \\
Qwen3.5-9B & - & 72.9 & {\bfseries 73.7} & 72.1 & +1.6 &  & {\bfseries 74.2} & 71.6 & +2.6 &  & {\bfseries 10.4} & 8.3 & 8.4 \\
Qwen3.5-27B & - & 83.7 & {\bfseries 85.4} & 81.9 & +3.5 &  & {\bfseries 88.5} & 78.9 & +9.6 &  & {\bfseries 10.2} & 2.8 & 3.3 \\
\hline
\tabH Qwen3.5-0.8B & \checkmark & 24.2 & 23.2 & {\bfseries 25.1} & -1.9 &  & 24.1 & {\bfseries 24.3} & -0.3 &  & {\bfseries 26.0} & 25.1 & 24.7 \\
Qwen3.5-2B & \checkmark & 60.7 & 58.2 & {\bfseries 63.0} & -4.8 &  & 60.6 & {\bfseries 60.7} & -0.1 &  & {\bfseries 19.0} & 12.6 & 7.7 \\
Qwen3.5-4B & \checkmark & 89.9 & 89.0 & {\bfseries 90.8} & -1.8 &  & 89.9 & {\bfseries 90.0} & -0.1 &  & 3.6 & {\bfseries 4.3} & 2.2 \\
Qwen3.5-9B & \checkmark & 91.8 & 90.8 & {\bfseries 92.8} & -1.9 &  & 91.5 & {\bfseries 92.2} & -0.7 &  & 2.8 & {\bfseries 4.0} & 1.4 \\
Qwen3.5-27B & \checkmark & 97.3 & 96.9 & {\bfseries 97.7} & -0.8 &  & {\bfseries 97.5} & 97.1 & +0.4 &  & 1.1 & {\bfseries 1.2} & 0.4 \\
\bhline
\end{tabular}
\caption{Breakdown of Japanese FrameBench performance by sentence-pair type, position of the correct sentence, and error type.
$\mathcal{R}$ denotes reasoning mode, and Acc. denotes overall accuracy.
Base and Ext. report accuracy on base and extended sentence pairs, while 1st and 2nd report accuracy when the correct sentence appears first or second.
$\Delta_{B-E}$ and $\Delta_{1-2}$ indicate performance differences. 
Both denotes selecting both sentences, Neither denotes selecting neither sentence, and Opposite denotes selecting the incorrect sentence.
Acc. and the three error-type rates sum to 100\% up to rounding.}
\label{tab:discussion-ja}
\end{table*}
These results suggest that the overall performance trends are largely preserved when a different LLM is used in dataset construction.

\section{Analysis on the Japanese Subset}
\label{app:ja_analysis}
\after{Table~\ref{tab:discussion-ja} reports the behavioral analysis on the Japanese subset of FrameBench, covering sentence-pair type, correct sentence position, and error type.
The Japanese results are generally consistent with the English analysis for correct sentence position and error type, but differ in sentence-pair type.
In English, extended pairs show the expected advantage over base pairs, with a gap of 2.57 points in human scores.
In Japanese, this effect is weaker: the corresponding gap decreases to 0.93 points, suggesting that the surface-level contrast between base and extended pairs is less pronounced.
Consistent with this weaker contrast, the LLM results in Japanese show a less consistent advantage for extended pairs.}

\end{document}